\documentclass{article}

\usepackage{microtype}
\usepackage{graphicx}
\usepackage{subcaption}
\usepackage{booktabs}

\usepackage{amsmath}
\usepackage{amssymb}
\usepackage{mathtools}
\usepackage{amsthm}

\usepackage{hyperref}

\usepackage[accepted]{icml2026}

\usepackage[capitalize,noabbrev]{cleveref}

\usepackage{threeparttable}
\usepackage{xcolor}
\usepackage[most,breakable]{tcolorbox}
\usepackage{tabularx}
\usepackage{url}

\theoremstyle{plain}

\theoremstyle{definition}

\theoremstyle{remark}

\newcommand{\pvar}[1]{\texttt{\{#1\}}}

\definecolor{PromptBlue}{RGB}{35,90,140}
\definecolor{PromptGreen}{RGB}{40,110,70}
\definecolor{PromptRed}{RGB}{130,50,50}
\definecolor{PromptGray}{RGB}{245,245,245}

\newtcolorbox{promptbox}[2][]{
  breakable,
  enhanced,
  colback=PromptGray,
  colframe=#2,
  boxrule=0.8pt,
  left=6pt,right=6pt,top=6pt,bottom=6pt,
  fonttitle=\bfseries,
  title={#1}
}

\icmltitlerunning{Structured Evidence Routing for EHR Risk Prediction}

\begin{document}

\twocolumn[
  \icmltitle{Structured Evidence Routing for Incident Risk Prediction from Multimodal Longitudinal EHRs}

  \begin{icmlauthorlist}
    \icmlauthor{Animesh Agarwal}{optum}
    \icmlauthor{Meysam Ghaffari}{optum}
    \icmlauthor{Nina Fatehi}{optum}
    \icmlauthor{Carlos Morato}{optum}
  \end{icmlauthorlist}

  \icmlaffiliation{optum}{Optum AI, UnitedHealth Group, Minneapolis, Minnesota, USA}

  \icmlcorrespondingauthor{Animesh Agarwal}{animesh.agarwal@optum.com}

  \icmlkeywords{structured health data, electronic health records, risk stratification, multimodal learning, interpretability}

  \vskip 0.3in
]

\printAffiliationsAndNotice{}

\begin{abstract}
Incident risk prediction from longitudinal electronic health records (EHRs) is
challenging because relevant signals are multimodal, weak in isolation, and
distributed across irregular patient histories. We propose structured evidence
routing, a router--predictor--reviewer workflow that separates full-record
access from disease-specific assessment. The router organizes the complete
pre-index EHR into a compact summary and targeted evidence slices; the
predictor uses this evidence to form an evidence-linked risk assessment, which
the reviewer critiques. For comparison with supervised EHRSHOT baselines, we
pair the routed evidence summaries with a supervised classifier readout. Across
five 1-year incident diagnosis tasks, our method reaches the AUROC range of
established supervised EHRSHOT baselines and remains competitive on AUPRC,
while exposing a patient-specific evidence trail. Internal pre-readout
ablations further suggest that routing, laboratory evidence, task guidance, and
review each contribute to performance.
\end{abstract}

\section{Introduction}

Many clinically important conditions emerge gradually, with early warning signals distributed across years of care. The practical challenge is therefore not only to diagnose disease once it is obvious, but to identify patients who are on track to develop a new condition while intervention is still possible. In structured longitudinal electronic health records (EHRs), the relevant signal is rarely concentrated in one modality: clinicians integrate diagnoses, medications, laboratory trajectories, physiologic measurements, procedures, and utilization patterns over time when assessing future risk. The challenge is thus not only prediction from tabular or sequential data, but prediction from \emph{heterogeneous, irregular, multimodal structured health records}.

A large literature has improved prediction from longitudinal structured EHRs using recurrent, transformer-based, and pretrained foundation-style models, including methods for future diagnosis prediction and related forecasting tasks \citep{choi2016doctorai, choi2016retain, ma2017dipole, li2020behrt, li2023hibehrt, rasmy2021medbert, yang2023transformehr, steinberg2024motor}. These approaches have substantially advanced representation learning for structured patient histories, but incident risk prediction remains difficult when signals are weak, temporally diffuse, and distributed across modalities. In practice, many pipelines still rely primarily on coded diagnoses, procedures, and medications, while laboratory and physiologic observations are not always incorporated in an equally rich or explicit way. More recent work has explored retrieval augmentation and prompting for clinical prediction from structured histories \citep{xu2024ramehr, sho2024, renc2024ethos}. EHRSHOT provides a standardized benchmark for structured EHR prediction \citep{wornow2023ehrshot}, and recent work uses it to compare count-based, pretrained sequential, and inference-time prompting pipelines under shared protocols \citep{gao2025countbased}.

This paper studies evidence organization for incident prediction from
multimodal longitudinal EHRs. Passing a full chart directly to a predictor can
bury weak disease-specific signals, while narrow feature views can miss repeated
abnormal measurements, treatment changes, or cross-visit trends. We propose a
router--predictor--reviewer workflow that keeps full-record access separate
from disease-specific assessment: the router summarizes the pre-index record
and returns targeted evidence slices, the predictor uses task guidance to score
risk, and the reviewer evaluates the intermediate assessment for
unsupported claims, missing evidence, or temporal inconsistencies.

We evaluate on five 1-year incident diagnosis tasks from EHRSHOT: hypertension,
hyperlipidemia, lupus, acute myocardial infarction, and pancreatic cancer. For
the main benchmark comparison, we pair routed evidence summaries with a
supervised classifier readout, matching the task-label access used by
count-based models, CLMBR, and MoA+Cls. We also report within-framework
pre-readout ablations to assess the contribution of routing, laboratory inputs,
task guidance, and review.

Our contributions are threefold: (i) we use the full pre-index structured EHR
for incident risk prediction, including diagnoses, procedures, medications,
laboratory results, measurements, observations, and utilization history; (ii) we
introduce structured evidence routing to separate full-record access from
disease-specific assessment while producing patient-specific evidence traces;
and (iii) we evaluate routed summaries as supervised prediction representations
on five EHRSHOT tasks, with comparisons to count-based, CLMBR, and MoA+Cls
baselines and pre-readout ablations of routing, laboratory evidence, task
guidance, and review.
\begin{figure*}[t]
    \centering
    \includegraphics[width=\textwidth]{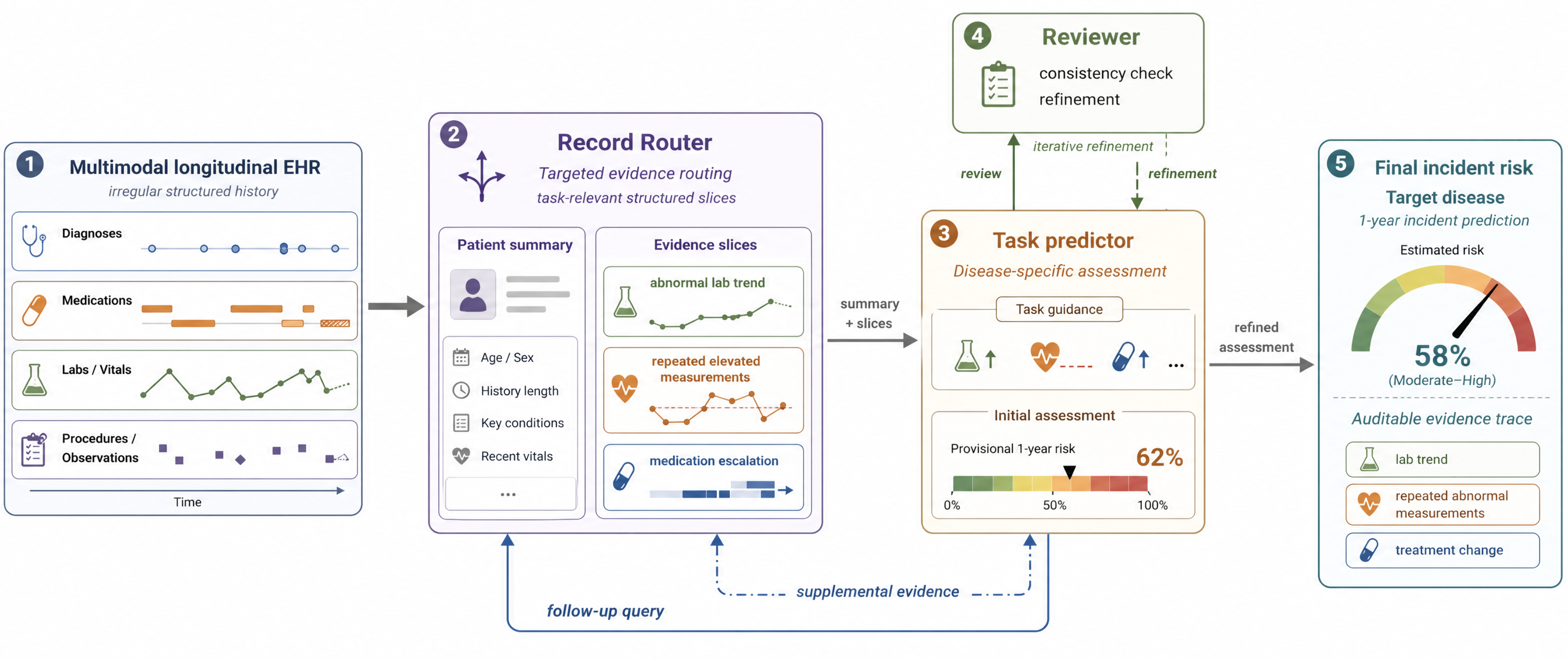}
\caption{Structured evidence routing for incident risk prediction. A record router organizes long, irregular EHR histories into a compact summary and targeted evidence slices. A review step can trigger refinement before final disease-specific prediction.}
\label{fig:overview}
\end{figure*}
\section{Method}

We now define the incident prediction setting and describe the three components
shown in Figure~\ref{fig:overview}: the Record Router, Task Predictor, and
Reviewer. The design is related to gatekeeper-style diagnostic orchestration,
but targets structured longitudinal EHR prediction rather than sequential
diagnosis \citep{sequentialdx2025}.

\subsection{Problem Setup}

For each patient, let $X^{(\leq t^\ast)} = \{e_t \mid t \leq t^\ast\}$ denote the structured longitudinal EHR history observed up to prediction time $t^\ast$, where each event may include diagnoses, medications, procedures, laboratory results, measurements, and other structured observations. Given a target disease $y$ and prediction horizon $h$, the task is to estimate
\begin{equation}
\hat{p}_{y,h} \approx P\!\left(Y_y(t^\ast + h)=1 \mid X^{(\leq t^\ast)}\right),
\end{equation}
that is, the probability that the patient will develop an incident outcome within the future horizon using only information available before the prediction time.

The framework operates directly on the full pre-index longitudinal EHR rather
than on a fixed hand-engineered feature vector. For each benchmark instance,
the available record includes structured events observed before the
task-specific prediction time across diagnoses, procedures, medications,
laboratory results, measurements, and observations.
All three inference modules are implemented as prompted LLM components, but
they have different access privileges and roles.

The \textbf{Record Router} is the only module allowed to access the full
structured record. It is constrained to organize evidence rather than make the
final disease-specific prediction. It serves two roles. First, it constructs a
compact patient summary containing salient active conditions, major medications,
recent utilization, and notable laboratory or physiologic trends. Second, it
returns \emph{targeted evidence slices} in response to disease-specific
follow-up requests, such as selected diagnosis history, laboratory trajectories,
medication changes, or time-localized observations.

The \textbf{Task Predictor} is a disease-specific LLM assessment module, with
task guidance derived from established clinical guidelines for the target
condition. It does not access the raw chart directly; instead, it receives the
router summary, identifies what additional evidence is needed under
task-specific clinical criteria, and requests targeted slices from the router.
Using the summary, retrieved evidence, and task guidance, the predictor produces
an initial incident-risk estimate with a short evidence-linked rationale.

The \textbf{Reviewer} is an LLM critique module that evaluates the intermediate
assessment for unsupported claims, missing evidence, or temporal
inconsistencies. When needed, it triggers iterative refinement: the predictor
requests additional evidence from the router, updates its assessment, and may
repeat this process until the reviewer finds the prediction sufficiently
supported or a preset iteration limit is reached. Figure~\ref{fig:overview}
illustrates the workflow. Prompt templates for the Record Router, Task Predictor, and Reviewer are provided in Appendix~\ref{app:prompt_templates}.

\section{Experimental Setup}

We evaluate on EHRSHOT in the 1-year incident diagnosis setting, following the
official task definitions and the latest-label protocol used by
\citet{gao2025countbased}. In this setting, each patient contributes at most
one prediction time per task: when multiple eligible labels are available, we
retain the most recent one. We use the official train/validation/test
partitioning from EHRSHOT \citep{wornow2023ehrshot} and evaluate five tasks:
hypertension, hyperlipidemia, lupus, acute myocardial infarction, and
pancreatic cancer. For each benchmark instance, only structured EHR events
observed on or before the prediction time are included.

For the main cross-method benchmark comparison, we report a supervised-readout
setting because the strongest structured-EHR baselines in this benchmark,
including count-based models and CLMBR, are trained on task labels. We first
run the structured evidence-routing framework to construct disease-specific
routed evidence summaries. These summaries are embedded with PubMedBERT
\citep{gu2021pubmedbert} and passed to an XGBoost classifier
\citep{chen2016xgboost} trained on benchmark training labels. This places our
method in the same task-label-access regime as count-based, CLMBR, and MoA+Cls
baselines.
This comparison treats each method as a supervised prediction pipeline built on
a different patient representation. Count-based models use coded event
features, CLMBR uses learned longitudinal EHR embeddings, MoA+Cls uses
LLM-generated summaries with a classifier head, and our method uses routed
evidence summaries with a classifier head. The comparison therefore asks whether
the evidence surfaced by the routing workflow can support risk scores in the
range of established supervised EHR baselines while preserving an auditable
evidence trail.

We additionally report internal ablations on the pre-readout framework. These
ablations are not intended as cross-method comparisons to EHRSHOT baselines.
Instead, they isolate the effect of removing components within the same
evidence-routing pipeline. We therefore do not train a separate supervised
classifier for each ablated variant; doing so would test whether a downstream
classifier can compensate for missing evidence-routing components, rather than
whether those components contribute to the framework's own prediction signal.
Cohort construction, preprocessing details, and supervised-readout details are
provided in Appendix~\ref{app:ehrshot_preprocessing} and
Appendix~\ref{app:calibration}. The clinical guideline sources used for task
guidance are summarized in Appendix~\ref{app:clinical_guidelines}.

\section{Results}

\subsection{Supervised Readout Comparison to Structured-EHR Baselines}

Tables~\ref{tab:ehrshot_cal_auroc} and~\ref{tab:ehrshot_cal_aupr} compare our
supervised-readout variant with established EHRSHOT baselines. The comparison
places routed evidence summaries alongside other supervised patient
representations: coded event features, learned longitudinal embeddings, and
LLM-generated summaries.

On AUROC, our supervised-readout variant is competitive across all five tasks
(Table~\ref{tab:ehrshot_cal_auroc}). It improves over CLMBR on every task and
matches or exceeds MoA+Cls on all five tasks, including gains on
hyperlipidemia, pancreatic cancer, and acute myocardial infarction. Compared
with the count-based baseline, our method matches performance on hypertension
and hyperlipidemia, exceeds it on lupus and acute myocardial infarction, and is
within $0.01$ AUROC on pancreatic cancer. These results indicate that routed
evidence summaries preserve enough longitudinal risk signal for effective
case ranking under a supervised readout.

AUPRC gives a complementary view of the results
(Table~\ref{tab:ehrshot_cal_aupr}), especially because incident diagnosis
tasks can be class-imbalanced. Under this metric, our method improves over
CLMBR on all five tasks, suggesting that routed evidence summaries provide a
stronger supervised prediction representation than learned longitudinal
embeddings in this setting. The method is also competitive with MoA+Cls:
it matches MoA+Cls on hypertension, exceeds it on hyperlipidemia and lupus,
and remains close on pancreatic cancer and acute myocardial infarction. Against
count-based models, which remain a strong supervised reference point, our
method reaches a similar range on multiple tasks and obtains the highest AUPRC
on lupus. Thus, the precision--recall results support the same overall
conclusion: structured evidence routing yields useful prediction features while
also exposing the disease-specific evidence behind the score.
\begin{table}[t]
\caption{\textbf{Supervised-readout AUROC on EHRSHOT} for 1-year incident diagnosis prediction. Our method uses routed evidence summaries with a trained classifier readout, placing it in the same supervised task-label setting as count-based, CLMBR, and MoA+Cls baselines. For Ours+Cls, $\pm$ denotes the bootstrap standard error estimated from
1{,}000 resamples; baseline uncertainties are reported as provided by prior
work.}
\label{tab:ehrshot_cal_auroc}
\centering
\small
\resizebox{\columnwidth}{!}{%
\begin{tabular}{lcccc}
\toprule
\textbf{Task} & \textbf{Count-based} & \textbf{CLMBR} & \textbf{MoA+Cls} & \textbf{Ours+Cls} \\
\midrule
Hypertension & $0.73 \pm 0.003$ & $0.70 \pm 0.003$ & $0.73 \pm 0.01$ & $0.73 \pm 0.03$ \\
Hyperlipidemia & $0.75$ & $0.70$ & $0.72 \pm 0.01$ & $0.75 \pm 0.02$ \\
Lupus & $0.76$ & $0.77 \pm 0.04$ & $0.82 \pm 0.03$ & $0.82 \pm 0.04$ \\
Pancreatic cancer & $0.89$ & $0.82$ & $0.84 \pm 0.005$ & $0.88 \pm 0.03$ \\
Acute myocardial infarction & $0.76 \pm 0.01$ & $0.74 \pm 0.01$ & $0.76 \pm 0.01$ & $0.77 \pm 0.02$ \\
\bottomrule
\end{tabular}%
}
\vspace{0.5ex}

{\footnotesize \textbf{MoA+Cls}: Qwen + BGE classifier + trained classifier head. \textbf{Ours+Cls}: structured evidence-routing summaries + PubMedBERT embeddings + trained classifier head.}
\end{table}

\begin{table}[t]
\centering
\caption{\textbf{Supervised-readout AUPRC on EHRSHOT} for 1-year incident diagnosis prediction. For Ours+Cls, $\pm$ denotes the bootstrap standard error estimated from
1{,}000 resamples; baseline uncertainties are reported as provided by prior
work.}
\label{tab:ehrshot_cal_aupr}
\small
\resizebox{\columnwidth}{!}{%
\begin{tabular}{lcccc}
\toprule
\textbf{Task} & \textbf{Count-based} & \textbf{CLMBR} & \textbf{MoA+Cls} & \textbf{Ours+Cls} \\
\midrule
Hypertension & $0.40 \pm 0.01$ & $0.32 \pm 0.01$ & $0.36 \pm 0.02$ & $0.36 \pm 0.03$ \\
Hyperlipidemia & $0.36 \pm 0.005$ & $0.29$ & $0.30 \pm 0.01$ & $0.34 \pm 0.04$ \\
Lupus & $0.14 \pm 0.04$ & $0.10 \pm 0.01$ & $0.16 \pm 0.03$ & $0.18 \pm 0.07$ \\
Pancreatic cancer & $0.39 \pm 0.04$ & $0.20$ & $0.36 \pm 0.01$ & $0.34 \pm 0.07$ \\
Acute myocardial infarction & $0.27 \pm 0.01$ & $0.23 \pm 0.01$ & $0.28 \pm 0.03$ & $0.27 \pm 0.04$ \\
\bottomrule
\end{tabular}%
}
\vspace{0.5ex}

{\footnotesize 
\textbf{MoA+Cls}: Qwen + BGE classifier + trained classifier head. 
\textbf{Ours+Cls}: structured evidence-routing summaries + PubMedBERT embeddings + trained classifier head.}
\end{table}

\subsection{Internal Ablations of the Evidence-Routing Framework}

These ablations are within-framework diagnostics: they evaluate whether
removing individual components weakens the framework's direct prediction signal
before the supervised readout.

We run targeted ablations on the EHRSHOT pancreatic cancer task, a
low-prevalence and clinically heterogeneous endpoint where relevant evidence may
be distributed across laboratory results, observations, procedures, and
utilization history. We remove the Reviewer, laboratory and observation inputs,
the Record Router, and disease-specific task guidance, while keeping the rest
of the framework fixed. Table~\ref{tab:ablation_ehrshot} reports pre-readout
AUROC for these within-framework comparisons.

The full framework achieves an AUROC of $0.86 \pm 0.03$. Removing laboratory
measurements or bypassing the Record Router produces the largest drops, reducing
AUROC to $0.83$. Removing review-based refinement or task-specific guidance
also lowers AUROC to $0.84$. These results suggest that the framework's direct
prediction signal is not driven by a single module alone, but by the combination
of multimodal structured inputs, mediated access to the longitudinal record,
task-specific assessment criteria, and iterative review. Additional details on
the ablation setup are provided in Appendix~\ref{app:ablations}.

\begin{table}[t]
\centering
\caption{\textbf{Internal pre-readout ablation} on pancreatic cancer.}
\label{tab:ablation_ehrshot}
\small
\begin{tabular}{lc}
\toprule
\textbf{Configuration} & \textbf{Pre-readout AUROC} \\
\midrule
Full framework                  & $0.86 \pm 0.03$ \\
$-$Reviewer                     & $0.84 \pm 0.033$ \\
$-$Lab measurements              & $0.83 \pm 0.034$ \\
$-$Record Router                 & $0.83 \pm 0.037$ \\
$-$Task guidance                 & $0.84 \pm 0.030$ \\
\bottomrule
\end{tabular}
\end{table}

\section{Discussion}

Across five EHRSHOT tasks, the supervised-readout results show that routed
evidence summaries are effective patient representations for incident risk
prediction. They improve over CLMBR across tasks, remain competitive with
MoA+Cls, and reach the range of strong supervised baselines under both AUROC
and AUPRC. These results suggest that explicitly organizing multimodal
longitudinal evidence before readout can preserve clinically useful risk signal.

The value of the framework is not only predictive performance, but the form of
the intermediate representation. Unlike count-based features or pretrained
longitudinal embeddings, the proposed workflow surfaces a compact patient
summary, targeted disease-relevant evidence slices, and an evidence-linked
representation before scoring. This makes the path from longitudinal record to
risk score more explicit: the readout operates on inspectable evidence rather
than an opaque feature vector alone. The ablations further show that routing,
laboratory evidence, task guidance, and review each contribute to the direct
prediction signal, supporting the claim that evidence selection and organization
matter for multimodal longitudinal risk prediction.
\section{Limitations}

This study has clinical-evaluation limitations. Evaluation is
retrospective and based on benchmark labels, so the results do not yet measure
how clinicians would interpret or use the surfaced evidence in practice. In
particular, we do not evaluate whether the evidence traces improve clinician
trust, reduce review burden, or support earlier recognition of incident disease.
A natural next step is clinician-facing evaluation of the routed evidence,
rationales, and risk estimates in realistic review workflows.

\newpage
\appendix
\onecolumn
\providecommand{\pvar}[1]{\texttt{\{#1\}}}

\section{EHRSHOT Ablation Details}
\label{app:ablations}

The ablation study in Table~\ref{tab:ablation_ehrshot} is performed on the
EHRSHOT pancreatic cancer task before the supervised classifier readout. Each
variant removes one component while keeping the rest of the framework fixed:
removing the Reviewer stops the pipeline after the initial prediction round;
removing laboratory inputs excludes lab-derived measurements and observations
from the patient record; removing the Record Router exposes the raw structured
EHR directly to the predictor; and removing task guidance omits
guideline-derived disease-specific instructions from the predictor prompt. These
ablations isolate how routing, multimodal evidence, task guidance, and review
affect the framework's direct prediction signal without training a new readout
for each variant.

\section{Prompt Templates for EHRSHOT}
\label{app:prompt_templates}

This appendix summarizes the prompt templates used by the three inference modules in the EHRSHOT experiments: the \textbf{Record Router}, the \textbf{Task Predictor}, and the \textbf{Reviewer}. Variable placeholders are shown as \pvar{variable\_name}. Disease-specific task guidance is injected at runtime as \pvar{task\_guidance}.

\subsection{System Overview}

The system uses three modules in a structured evidence-routing loop. The \textbf{Record Router} is the only module with access to the longitudinal EHR and is responsible for summarization and targeted evidence retrieval. The \textbf{Task Predictor} is a disease-specific module that requests evidence, forms an initial risk assessment, and updates that assessment after review. The \textbf{Reviewer} evaluates whether the intermediate assessment is evidence-grounded and task-aligned, and can trigger an additional evidence request.

\subsection{Record Router Prompts}

\subsubsection*{Prompt R-1: Patient Summarization}

\begin{quote}\small
\textbf{System:} You are a clinical record router with access to a patient's longitudinal structured EHR.

\textbf{Task:} Produce a concise, neutral summary of the patient record for downstream prediction.

\textbf{Instructions:}
\begin{itemize}
  \item Summarize demographics and major recent or recurrent clinical history.
  \item Highlight high-signal diagnoses, medications, laboratory findings, and observations relevant to future disease risk.
  \item Use plain clinical language and avoid unsupported inference.
  \item Do not diagnose, speculate, or recommend treatment.
\end{itemize}

\textbf{Input:} \pvar{patient\_metadata}, \pvar{clinical\_data}, \pvar{family\_history}

\textbf{Output:} A brief neutral paragraph summarizing the patient context.
\end{quote}

\subsubsection*{Prompt R-2: Targeted Evidence Retrieval}

\begin{quote}\small
\textbf{System:} You are a clinical record router responding to predictor questions using the structured patient record.

\textbf{Task:} For each question, retrieve only directly relevant evidence from the EHR.

\textbf{Instructions:}
\begin{itemize}
  \item Search diagnoses, medications, procedures, observations, laboratory results, and family history.
  \item Return matching evidence with dates and values when available.
  \item Prefer exact evidence over interpretation.
  \item If no relevant evidence is present, state that no relevant data were found.
\end{itemize}

\textbf{Input:} \pvar{patient\_metadata}, \pvar{clinical\_data}, \pvar{family\_history}, \pvar{predictor\_questions}

\textbf{Output:} Structured responses pairing each question with the retrieved evidence.
\end{quote}

\subsection{Task Predictor Prompts}

\subsubsection*{Prompt P-1: Question Generation}

\begin{quote}\small
\textbf{System:} You are a \pvar{disease} predictor. Your assessment should follow the supplied task guidance.

\textbf{Task:} Generate a small number of targeted questions for the Record Router to retrieve the evidence needed to assess 12-month incident risk of \pvar{disease}.

\textbf{Instructions:}
\begin{itemize}
  \item Prioritize primary diagnostic measurements, key symptoms, and major risk factors.
  \item Ask only questions that can be answered from the structured EHR.
  \item Do not score risk or interpret evidence at this stage.
  \item Avoid repeating questions from prior rounds.
\end{itemize}

\textbf{Input:} \pvar{patient\_summary}, \pvar{prior\_questions}, [\pvar{reviewer\_feedback}], \pvar{task\_guidance}

\textbf{Output:} A short list of targeted evidence requests.
\end{quote}

\subsubsection*{Prompt P-2: Risk Assessment}

\begin{quote}\small
\textbf{System:} You are a \pvar{disease} predictor. Use only the supplied task guidance and the Record Router-provided evidence.

\textbf{Task:} Assess the probability that the patient will receive a \emph{new} \pvar{disease} diagnosis within 1 year after the index date.

\textbf{Instructions:}
\begin{itemize}
  \item Ground the assessment in task-relevant thresholds, recency, and trends.
  \item Distinguish incident risk from pre-existing disease.
  \item Treat missing tests as missing evidence, not negative findings.
  \item When evidence is limited, state that uncertainty explicitly.
\end{itemize}

\textbf{Input:} \pvar{patient\_summary}, \pvar{router\_responses}, \pvar{task\_guidance}

\textbf{Output:} A structured risk profile containing a risk score, key supporting evidence, countervailing evidence, and a short rationale.
\end{quote}

\subsubsection*{Prompt P-3: Post-Review Update}

\begin{quote}\small
\textbf{Task:} Revise the predictor assessment after reviewing the Reviewer feedback.

\textbf{Instructions:}
\begin{itemize}
  \item Address hard errors and important soft concerns raised by the Reviewer.
  \item Ask additional questions only if new evidence is needed.
  \item Do not repeat previously asked questions.
  \item If no additional information is needed, update the assessment directly.
\end{itemize}

\textbf{Input:} \pvar{patient\_summary}, \pvar{reviewer\_feedback}, \pvar{prior\_questions}, \pvar{router\_responses}

\textbf{Output:} Either a short list of new questions or an updated structured risk profile.
\end{quote}

\subsection{Reviewer Prompts}

\subsubsection*{Prompt V-1: Review}

\begin{quote}\small
\textbf{System:} You are the Reviewer in a structured evidence-routing \pvar{disease} risk prediction system.

\textbf{Task:} Evaluate whether the current assessment is evidence-grounded, temporally appropriate, and aligned with the incident prediction task.

\textbf{Instructions:}
\begin{itemize}
  \item Identify unsupported claims, temporal mismatches, and incident-versus-prevalent errors.
  \item Recommend only small score adjustments unless a clear hard error is present.
  \item Prefer no change when the assessment is coherent and evidence-grounded.
\end{itemize}

\textbf{Input:} \pvar{current\_risk\_profile}, \pvar{qa\_history}

\textbf{Output:} A short structured review containing a summary, hard errors, soft concerns, and an optional score adjustment recommendation.
\end{quote}

\subsubsection*{Prompt V-2: Final Prediction Summary}

\begin{quote}\small
\textbf{Task:} Summarize the final disease-specific prediction after refinement has concluded.

\textbf{Instructions:}
\begin{itemize}
  \item Base the summary only on the final predictor risk profile.
  \item Report the final score for the active disease.
  \item Keep the rationale concise and evidence-based.
\end{itemize}

\textbf{Input:} \pvar{predictor\_profile}

\textbf{Output:} A short rationale summarizing the final prediction.
\end{quote}

\section{EHRSHOT Cohort Construction and Preprocessing}
\label{app:ehrshot_preprocessing}

This appendix summarizes the preprocessing steps used for the EHRSHOT evaluation. The goal was to reproduce the benchmark setting as closely as possible while converting the structured EHR stream into the longitudinal input format required by the structured evidence-routing pipeline.

\subsection{Benchmark Protocol and Test-Set Selection}

We follow the official EHRSHOT task definitions and patient-ID
train/validation/test splits \citep{wornow2023ehrshot}. We use the
latest-label evaluation setting of \citet{gao2025countbased}: because EHRSHOT
labels are assigned at the visit level, a patient may have multiple eligible
prediction times for the same task, and we retain only the most recent eligible
label per patient.

For held-out test evaluation, we load the official test patient IDs, filter the
task-specific labeled-patients file to positive and negative examples from that
split, and select the row with the latest prediction time for each patient. This
produces one benchmark test instance per patient per task. Training patients are used only for supervised-readout training and model
selection, as described in Appendix~\ref{app:calibration}; held-out test labels
are not used during readout training or selection.
\subsection{Temporal Cutoff and Longitudinal Input Construction}

For every retained patient, only EHR events observed on or before the benchmark prediction time were included in the model input.

The patient record was converted into a chronological encounter-style representation. Events were grouped by calendar date and organized into clinically interpretable modalities, including diagnoses, medications, procedures, measurements, observations, notes, visit details, and device exposures. Within each calendar date, duplicate same-day concept occurrences were collapsed so that the resulting record preserved temporal structure without redundant repetition.

The final output of preprocessing was a patient-level JSON record containing: (i) demographics, (ii) task metadata including prediction time and label, and (iii) a time-ordered list of pre-prediction encounters. This representation preserves the full structured longitudinal history available before the indexed benchmark time while remaining compatible with router-mediated summarization and targeted evidence retrieval.

\subsection{Code-to-Description Enrichment}

EHRSHOT represents clinical events using structured \texttt{<SYSTEM>/<CODE>} identifiers spanning SNOMED CT, LOINC, RxNorm, and CPT4. Before ingestion into the inference pipeline, these coded fields were enriched with human-readable descriptions using static terminology lookup tables, so that diagnoses, medications, procedures, laboratory tests, and observations were presented with both the original code and an interpretable descriptor.

\subsection{Preprocessing Choices Relevant to Interpretation}

Three preprocessing choices are especially important for interpreting the EHRSHOT results. First, we used the benchmark's official held-out test patients rather than constructing a new split, so the reported results are directly comparable to prior work following the same protocol. Second, we adopted the latest-label setting, meaning that each patient contributes at most one prediction time per task. Third, unlike count-based benchmark pipelines that operate on ontology roll-ups or restricted modality sets, the proposed method consumes the full pre-prediction structured event stream and relies on the Record Router to organize that evidence for downstream prediction.

\section{Supervised Readout over Routed Evidence}
\label{app:calibration}

The supervised-readout variant uses a classifier over the routed evidence
representation produced by the structured evidence-routing framework. This
readout is included because the primary EHRSHOT baselines used for comparison,
including count-based models and CLMBR, are also trained on task labels. The
goal is to compare patient representations under a shared supervised
task-adaptation setting.

For each patient in the EHRSHOT training split, we collected the final
framework output, including the disease-specific prediction summary, the final
round of task-specific reasoning, and the top evidence drivers identified by
the system. This information was condensed into a short clinical evidence
summary using a GPT-4.1-mini-based extractor. The summaries were embedded into
dense vector representations using PubMedBERT embeddings. A supervised
classifier head was then trained over these features using XGBoost, with model
selection performed by 5-fold stratified cross-validation on the training
split.

The selected classifier was applied to the held-out benchmark test set to
produce prediction scores for AUROC and AUPRC evaluation. Test labels were not
used during readout training or model selection. 

\section{Clinical Guideline Sources for Task Guidance}
\label{app:clinical_guidelines}

For each incident diagnosis task, the disease-specific predictor receives a
short task guidance prompt derived from established clinical guidelines summarized in Table~\ref{tab:guidelines}. These
references are used to define relevant risk factors, supporting evidence, and
temporal patterns for evidence retrieval and risk assessment.

\begin{center}
\footnotesize
\textbf{Table~\refstepcounter{table}\thetable. Clinical guideline and reference sources used for disease-specific task guidance.}
\label{tab:guidelines}

\vspace{0.5em}

\begin{tabular}{p{1.8cm} p{5.7cm}}
\toprule
\textbf{Condition} & \textbf{Guideline / reference} \\
\midrule
Hypertension &
2025 AHA/ACC Guideline for the Prevention, Detection, Evaluation, and
Management of High Blood Pressure in Adults~\cite{aha2025} \\[3pt]
Hyperlipidemia &
2026 AHA/ACC/Multisociety Guideline on the Management of
Dyslipidemia~\cite{aha2026dyslipidemia} \\[3pt]
Lupus &
2019 EULAR/ACR Classification Criteria for Systemic Lupus
Erythematosus~\cite{aringer2019sle} \\[3pt]
Acute MI &
Fourth Universal Definition of Myocardial Infarction
(2018)~\cite{thygesen2018mi} \\[3pt]
Pancreatic cancer &
ESMO Clinical Practice Guideline for Diagnosis, Treatment and
Follow-up~\cite{conroy2023pancreatic} \\
\bottomrule
\end{tabular}
\end{center}
\end{document}